\documentclass[twocolumn,a4paper,nofootinbib]{revtex4-1}
\usepackage{amsmath,amssymb,amsfonts}
\usepackage{graphicx}
\usepackage{xcolor}
\usepackage{braket}
\usepackage{physics}
\usepackage{subfigure}
\usepackage{algorithm}
\usepackage{algpseudocode}
\usepackage{bm}
\usepackage{hyperref}
\usepackage{textcomp}
\usepackage{url}

\begin{document}

\title{Tensor Network Generator-Enhanced Optimization for Traveling Salesman Problem}

\author{Ryo Sakai}
\email{r.sakai@j-ij.com}
\affiliation{JIJ Inc., 3-3-6 Shibaura, Minato-ku, Tokyo, 108-0023, Japan}

\author{Chen-Yu Liu}
\email{d10245003@g.ntu.edu.tw}
\affiliation{Graduate Institute of Applied Physics, National Taiwan University, Taipei, Taiwan}

\begin{abstract}
  We present an application of the tensor network generator-enhanced optimization (TN-GEO) framework to address the traveling salesman problem (TSP), a fundamental combinatorial optimization challenge.
  Our approach employs a tensor network Born machine based on automatically differentiable matrix product states (MPS) as the generative model, using the Born rule to define probability distributions over candidate solutions.
  Unlike approaches based on binary encoding, which require $N^2$ variables and penalty terms to enforce valid tour constraints, we adopt a permutation-based formulation with integer variables and use autoregressive sampling with masking to guarantee that every generated sample is a valid tour by construction.
  We also introduce a $k$-site MPS variant that learns distributions over $k$-grams (consecutive city subsequences) using a sliding window approach, enabling parameter-efficient modeling for larger instances.
  Experimental validation on TSPLIB benchmark instances with up to 52 cities demonstrates that TN-GEO can outperform classical heuristics including swap and 2-opt hill-climbing.
  The $k$-site variants, which put more focus on local correlations, show better results compared to the full-MPS case.
\end{abstract}


\maketitle

\section{Introduction}

Quantum computing has emerged as a transformative technology, offering significant potential to enhance computational performance across various domains, particularly in combinatorial optimization~\cite{Lucas:2013ahy,Abbas:2023agz}.
Among these challenges, the traveling salesman problem (TSP) stands out due to its conceptual simplicity and computational complexity.
Given a set of $N$ cities with pairwise distances $d_{ij}$, the TSP asks for a permutation $\mathbf{p} = (p_1, p_2, \ldots, p_N)$ of the city indices that minimizes the total tour length
\begin{align}
  D(\mathbf{p}) = \sum_{i=1}^{N-1} d_{p_i, p_{i+1}} + d_{p_N, p_1}.
\end{align}
The problem is NP-hard~\cite{Lucas:2013ahy}, and the number of feasible solutions, $(N-1)!/2$ for symmetric instances, grows factorially with the number of cities.
By leveraging unique quantum phenomena such as superposition and entanglement, quantum computing holds the promise of addressing such combinatorial explosions more effectively than traditional classical methods.

The current era of noisy intermediate-scale quantum (NISQ)~\cite{Bharti:2021zez} technologies faces significant challenges, particularly due to the constraints imposed by the limited number of qubits available on quantum devices.
Quantum-inspired approaches~\cite{Arrazola:2019kwz,Tang:2018lmw}, in particular those based on tensor networks, offer tractable classical implementations that retain the mathematical framework of quantum mechanics---such as Born rule probabilities and wave function amplitudes---while providing tunable expressiveness through the bond dimension, which controls the degree of entanglement the model can represent.
Among these, tensor network generator-enhanced optimization (TN-GEO)~\cite{Alcazar:2021wjc} has emerged as a promising framework, initially developed to address portfolio optimization problems within the realm of binary optimization.
Practical aspects of GEO~\footnote{
  In the original paper, it is stated that any generative model (not limited to tensor networks) can be used in the GEO framework,
  so GEO is a wider concept compared to TN-GEO.
  In this paper, however, we do not notationally distinguish TN-GEO from GEO in a strict sense.
} have been analyzed in ref.~\cite{Gardiner:2024fzq};
the generalization capabilities of the underlying tensor network Born machine---namely, the ability to generate unseen valid samples with lower cost than the training data---have been formally characterized in ref.~\cite{Gili:2022oul};
and the framework has been successfully applied to various problems including the multi-knapsack problem~\cite{Vodovozova:2025idx} and industrial-scale production planning~\cite{Banner:2023abq}.
This approach exemplifies the potential of quantum-inspired techniques to tackle complex computational challenges, even in the absence of fully developed quantum hardware.

Most quantum and quantum-inspired TSP solvers~\cite{10.1007/s00500-013-1203-7,10.3389/fphy.2021.760783} use binary encoding with $N^2$ variables.
This means embedding a solution space that is originally of $N!$ permutations into a vastly expanded space of $2^{N^{2}}$, the majority of which consists of invalid states that do not satisfy the constraints.
Penalty terms are then required to steer the optimization toward feasible solutions, which can create a rugged energy landscape.
Strategies to mitigate the $N^2$ scaling include path-slicing approaches that decompose the tour into smaller subproblems~\cite{Liu:2024eoi}.
An alternative is integer encoding, which avoids the $N^2$ expansion by construction and has been explored within GEO for the knapsack problem~\cite{Vodovozova:2025idx}.
For TSP, however, an additional mechanism is needed to enforce the permutation constraint---simply using integer variables does not prevent a city from being visited more than once.
More broadly, learning and generating permutations is a known challenge for generative models~\cite{NIPS2015_29921001,mena2018learning}.

As in the original TN-GEO work~\cite{Alcazar:2021wjc}, we employ a matrix product state (MPS) Born machine~\cite{Han:2018jah} as the generative model, but represent tours as sequences of $N$ integer-valued variables, with the MPS directly modeling distributions over city indices.
To enforce valid permutations, we employ an autoregressive sampling procedure with masking: at each position, the model samples from the conditional distribution over remaining unvisited cities, guaranteeing that every generated sample is a valid tour without rejection sampling or post-selection.
This choice is particularly well suited to MPS-based models: with $N$ integer sites, consecutive positions in the MPS chain correspond to consecutive cities in the tour, so the local correlations that MPS captures with finite bond dimension align directly with the local structure of TSP solutions.
In contrast, a binary encoding would inflate the chain to $N^2$ sites, diluting these local correlations and forcing the model to spend capacity learning to suppress the exponentially many invalid configurations.
For constrained sampling in tensor networks, related approaches include symmetry-embedded MPS~\cite{Lopez-Piqueres:2022adw,Lopez-Piqueres:2025vbd}, constrained MPS architectures~\cite{Nakada:2024coe}, and projection layers that filter out invalid permutations~\cite{MataAli:2023wlv}.

The alignment between MPS chain structure and tour sequence raises a natural question: how much of the tour does the model need to capture at once?
We investigate this through a $k$-site MPS variant that learns distributions over $k$-grams---consecutive subsequences of $k$ cities---rather than entire tours.
Our experimental results reveal that moderate values of $k$ (4--8) consistently outperform both smaller ($k=2$) and full ($k=N$) models, suggesting that near-optimal TSP tours are largely determined by local geometric patterns among consecutive cities.

The remainder of this paper is organized as follows.
Section~\ref{sec:method} describes the GEO iterative optimization framework and the tensor network Born machine.
Section~\ref{sec:result} presents experimental results on TSPLIB instances.
Section~\ref{sec:conclusion} concludes with a discussion of future directions.

\section{Method}
\label{sec:method}

We first review the overall GEO iterative optimization scheme, then describe the Born machine generative model and their key components.

\subsection{Generator-Enhanced Optimization framework}

\begin{figure*}[t]
  \includegraphics[width=2\columnwidth]{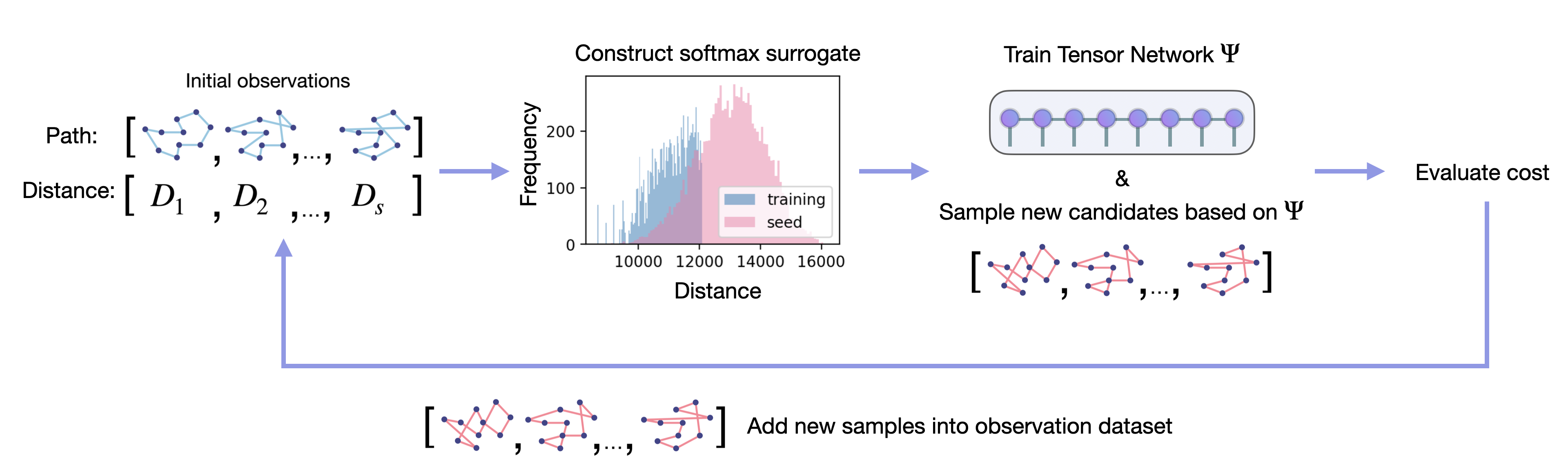}
  \caption{
    Schematic picture of GEO for TSP.
    The framework iteratively refines a generative model by training on a softmax-weighted distribution of candidate solutions.
    At each iteration, the model generates new samples that are biased toward lower-cost tours.
  }
  \label{fig:scheme}
\end{figure*}

GEO~\cite{Alcazar:2021wjc} is an iterative framework that uses generative models to solve combinatorial optimization problems.
The key idea is to train a generative model to produce candidate solutions that are likely to have low cost, then use these samples to further refine the model.
The framework operates as illustrated in fig.~\ref{fig:scheme}.
We summarize the workflow in algorithm~\ref{alg:geo} for reproducibility.
Note that, in this simplified presentation, every generated sample is evaluated by the cost function, omitting a weighting technique introduced in the original GEO framework~\cite{Alcazar:2021wjc} that is unnecessary for TSP (see below).

\begin{algorithm}[H]
  \caption{Generator-Enhanced Optimization~\cite{Alcazar:2021wjc}}
  \label{alg:geo}
  \begin{algorithmic}[1]
    \Require Cost function $C$, generative model $\mathcal{G}$, temperature $T$, number of iterations $n_{\mathrm{iters}}$
    \Ensure Best solution found
    \State Generate initial population $\mathcal{P}$ of candidate solutions
    \State Evaluate cost $C(\mathbf{x})$ for each $\mathbf{x} \in \mathcal{P}$
    \State $\mathbf{x}^{*} \gets \arg\min_{\mathbf{x} \in \mathcal{P}} C(\mathbf{x})$
    \For{$t = 1$ to $n_{\mathrm{iters}}$}
      \State Construct finite-temperature softmax distribution: $Q(\mathbf{x}) \propto \exp(-C(\mathbf{x})/T)$
      \State Train $\mathcal{G}$ to approximate $Q$ by minimizing negative log-likelihood
      \State Sample new candidates $\mathcal{S}$ from trained $\mathcal{G}$
      \State Evaluate cost $C(\mathbf{x})$ for each $\mathbf{x} \in \mathcal{S}$
      \State $\mathcal{P} \gets \mathcal{P} \cup \mathcal{S}$
      \If{$\min_{\mathbf{x} \in \mathcal{S}} C(\mathbf{x}) < C(\mathbf{x}^*)$}
        \State $\mathbf{x}^* \gets \arg\min_{\mathbf{x} \in \mathcal{S}} C(\mathbf{x})$
      \EndIf
    \EndFor
    \State \Return $\mathbf{x}^{*}$
  \end{algorithmic}
\end{algorithm}

This iterative refinement progressively focuses the search on promising regions of the solution space, with the generative model learning increasingly refined representations of good solutions.
This loop shares a conceptual similarity with reinforcement learning: the generative model plays the role of a policy that produces candidate solutions, the cost function provides a reward signal, and the softmax surrogate constructs an improved target distribution from which the policy is updated.

Note that, in the original paper, it is claimed that GEO can be used to boost the performance of other solvers.
In our current study, however, we exclusively use GEO as a standalone solver, meaning that we always start from a randomly sampled initial distribution.
Furthermore, in the original standalone mode, a key design goal was to minimize the number of cost function evaluations, motivated by applications where the objective function is expensive to evaluate (\textit{e.g.}, training a neural network or running a physical simulation).
For TSP, however, evaluating the tour length is simply a summation of pairwise distances, so it is computationally inexpensive compared to the MPS training and sampling steps.
We therefore evaluate the cost for every generated sample, as there is no practical benefit to reducing the number of cost evaluations in this setting.

\subsection{Softmax surrogate with temperature scheduling}

For optimization within the GEO framework, we train the generative model to match a softmax surrogate distribution constructed from the observed costs.
Given a dataset of tours $\{\mathbf{p}_{(i)}\}$ with corresponding lengths $\{D_{(i)}\}$, the target distribution is
\begin{align}
  Q_{i} = \frac{e^{-D_{(i)}/T}}{\sum_{j} e^{-D_{(j)}/T}},
\end{align}
where $T$ is a temperature hyperparameter.
Lower temperatures concentrate probability mass on lower-cost solutions, creating a sharper distribution that emphasizes the best solutions in the current population.

The model parameters (tensor elements) are optimized to minimize the negative log-likelihood
\begin{align}
  \mathcal{L} = -\sum_{i} Q_{i} \log P(\mathbf{p}_{(i)}),
\end{align}
where $P(\mathbf{p}_{(i)})$ is the model probability of observing $\mathbf{p}_{(i)}$.
This training objective encourages the model to assign higher probability to tours with shorter distances.

In the original GEO paper~\cite{Alcazar:2021wjc}, the temperature was held constant throughout optimization.
Here, to balance exploration and exploitation across iterations, we employ an exponential temperature schedule
\begin{align}
  T(t) = T_{\mathrm{init}} \left(\frac{T_{\mathrm{final}}}{T_{\mathrm{init}}}\right)^{(t-1)/(n_{\mathrm{iters}}-1)},
\end{align}
where $t$ is the current iteration ($1$-based) and where $T_{\mathrm{init}} > T_{\mathrm{final}}$ are the initial and final temperatures.
Early iterations use higher temperatures, allowing the model to learn from a broader distribution of solutions (exploration).
As iterations progress, the temperature decreases, causing the softmax distribution to concentrate more sharply on the best solutions found so far (exploitation)~\footnote{
  This cooling schedule may appear similar to simulated annealing, but the role of temperature differs between the two.
  In simulated annealing, the temperature controls the acceptance probability of cost-increasing moves during a Markov chain walk;
  lowering it reduces the probability of accepting cost-increasing moves and confines the search to local neighborhoods.
  In GEO, the temperature shapes the target distribution from which the generative model learns:
  lowering it sharpens the training signal so that the model concentrates on higher-quality solutions.
  Note also that the softmax surrogate in GEO inherently requires a temperature parameter, so some form of temperature scheduling---including a constant temperature as a trivial special case---is always present.
}.

\subsection{Tensor Network Born Machine}

A tensor network Born machine is a generative model that uses tensor networks to represent probability distributions via the Born rule from quantum mechanics.
The key idea is to parameterize a wave function $\Psi(\mathbf{x})$ using a tensor network, and interpret the squared amplitude $|\Psi(\mathbf{x})|^{2}$ as the (unnormalized) probability of observing configuration $\mathbf{x}$.
Recent theoretical analysis shows that MPS-based models occupy a favorable regime between trainability and expressiveness, avoiding the barren plateaus that affect more complex quantum circuits~\cite{Herbst:2025sgh}.

In this work, we employ the MPS to represent the wave function~\cite{Han:2018jah}~\footnote{
  Prior to ref.~\cite{Han:2018jah}, MPS was used for classification tasks in a pioneering work~\cite{stoudenmire2016supervised}.
}.
While the original TN-GEO framework operates on binary vectors encoding portfolio selections, a direct binary encoding for TSP would require $N^2$ variables, expanding the solution space to $2^{N^2}$ despite only $N!$ valid tours.
This means the majority of configurations represent invalid ones, typically handled through penalty terms~\cite{Lucas:2013ahy}.
Instead, we represent a TSP tour as a sequence of $N$ integer-valued variables $\mathbf{x} = (x_1, x_2, \ldots, x_N)$, where each $x_k \in \{0, 1, \ldots, N-1\}$ indicates which city is visited at position $k$.
Then the MPS is constructed with physical dimension $N$ at each site, allowing it to directly model distributions over city indices.
Note that the MPS itself can assign non-zero amplitude to invalid configurations;
the restriction to valid permutations is enforced during sampling through masking (see the next subsection for details).

The MPS wave function is given by
\begin{align}
  \Psi(\mathbf{x}) = \sum_{\{\alpha\}} A^{[1]}_{x_1, \alpha_1} A^{[2]}_{\alpha_1, x_2, \alpha_2} \cdots A^{[N]}_{\alpha_{N-1}, x_N},
\end{align}
where $A^{[k]}$ are rank-3 tensors (rank-2 for boundary tensors) and $\alpha_k$ are virtual bond indices with dimension $\chi$.
The bond dimension $\chi$ controls the expressiveness of the MPS: larger $\chi$ allows the model to capture more complex correlations in the data.

The probability of a configuration $\mathbf{x}$ is given by the Born rule
\begin{align}
  P(\mathbf{x}) = \frac{|\Psi(\mathbf{x})|^2}{Z},
\end{align}
where the normalization factor is defined by
\begin{align}
  Z = \sum_{\mathbf{x}'} |\Psi(\mathbf{x}')|^2.
\end{align}

A key advantage of Born machines over energy-based models such as Boltzmann machines~\cite{ACKLEY1985147} is that samples can be generated directly from the Born rule probabilities via perfect sampling~\cite{Ferris:2012pbh}, without requiring iterative procedures like Markov chain Monte Carlo or Gibbs sampling.
In the TSP case, however, we need additional techniques to sample only valid permutations, as explained in the next subsection.

\subsection{Autoregressive sampling of permutations}

A key advantage of the MPS structure is that it enables efficient autoregressive sampling.
By bringing the MPS into right-canonical form, the right environment becomes trivially an identity, so we can efficiently sample each variable sequentially from left to right, with the conditional probability of $x_k$ given previously sampled values computed exactly.

For TSP applications, we require samples to be valid permutations---each city must appear exactly once.
We achieve this through autoregressive sampling with masking: at each position $k$, we compute the conditional probabilities over all $N$ cities via the Born rule, mask out cities that have already been selected, and sample from the remaining valid options.
This ensures that every generated sample is a valid TSP tour without requiring rejection sampling or post-selection.
The sampling procedure is summarized in algorithm~\ref{alg:sampling}.

\begin{algorithm}[H]
  \caption{Autoregressive sampling with masking}
  \label{alg:sampling}
  \begin{algorithmic}[1]
    \Require Right-canonical MPS tensors $\{A^{[k]}\}_{k=1}^{N}$
    \Ensure Valid permutation $(x_{1}, \ldots, x_{N})$
    \State $L \gets (1, 0, \ldots, 0)^{\mathrm{T}} \in \mathbb{R}^{\chi}$ \Comment{Left environment vector}
    \State $\mathcal{U} \gets \emptyset$ \Comment{Set of visited cities}
    \For{$k = 1$ to $N$}
      \State $C_{j, \alpha} \gets \sum_{\beta} L_{\beta} A^{[k]}_{\beta, j, \alpha}$ for all $j, \alpha$ \Comment{Contract left environment}
      \State $p_j \gets \|C_{j}\|^2 = \sum_{\alpha} |C_{j, \alpha}|^2$ for all $j$ \Comment{Born rule probability (using the canonical property)}
      \State $p_j \gets 0$ for all $j \in \mathcal{U}$ \Comment{Mask used cities}
      \State $p_i \gets p_i / \sum_j p_j$ for all $i$ \Comment{Normalize}
      \State $x_k \gets$ sample from distribution $p$
      \State $L_{\alpha} \gets C_{x_k, \alpha}$ for all $\alpha$ \Comment{Update left environment}
      \State $\mathcal{U} \gets \mathcal{U} \cup \{x_k\}$
    \EndFor
    \State \Return $(x_1, \ldots, x_N)$
  \end{algorithmic}
\end{algorithm}

This approach shares the goal of pointer networks~\cite{NIPS2015_29921001}---generating valid permutations by sequentially selecting from remaining elements---but uses a fundamentally different mechanism: Born rule probabilities from tensor contractions rather than attention weights.
Learning and generating permutations is typically challenging for generative models~\cite{NIPS2015_29921001,mena2018learning}, but the MPS structure with masked autoregressive sampling naturally handles this constraint~\footnote{
  Alternative approaches construct MPS with built-in symmetry constraints that guarantee zero amplitude on invalid configurations~\cite{Lopez-Piqueres:2022adw,Lopez-Piqueres:2025vbd,Nakada:2024coe}.
}.

\subsection{GEO with $k$-site MPS}
\label{sec:k_site_geo}

The full $N$-site MPS described above uses one tensor per position in the tour, requiring $N$ tensors each with physical dimension $N$.
For larger problem instances, this can become computationally expensive.
Exploiting the TSP's symmetric problem structure, we introduce a more parameter-efficient variant: the $k$-site MPS, where $k < N$.

Instead of learning the joint distribution over entire tours, the $k$-site MPS learns distributions over $k$-grams---consecutive subsequences of $k$ cities.
For example, with $k=2$, the model learns edge distributions (pairs of consecutive cities); with $k=4$, it learns distributions over 4-city path segments.

Given full tours from the softmax surrogate distribution, we extract all overlapping $k$-grams as training data.
Thus a tour $(x_1, x_2, \ldots, x_N)$ yields $N$ training samples: $(x_1, \ldots, x_k)$, $(x_2, \ldots, x_{k+1})$, $\ldots$, $(x_{N-1}, \ldots, x_{k-2})$, $(x_{N}, \ldots, x_{k-1})$.

During sampling, we use a sliding window approach.
The model maintains a context of the most recent $k-1$ cities visited.
For the first $k-1$ positions, standard left-to-right contraction is used.
For subsequent positions, the left environment is rebuilt from the context window (last $k-1$ sampled cities) before contracting with the final tensor of the $k$-site MPS.
After sampling each city, the context window slides by dropping the oldest city and appending the newly sampled one.

The $k$-site MPS offers a parameter-efficiency trade-off: smaller $k$ requires fewer tensors ($k$ instead of $N$) but captures only local $k$-city correlations.
Larger $k$ captures longer-range dependencies but approaches the full model's complexity.
When $k = N$, this reduces to the full tour generation described earlier.
In the result section, we will show that intermediate values of $k$ can provide good solutions with reduced computational cost, especially for larger TSP instances.

\section{Results}
\label{sec:result}

\subsection{Experimental setup}
\label{sec:setup}

We evaluate TN-GEO on benchmark instances from TSPLIB~\cite{reinelt1991tsplib}, ranging from 14 to 52 cities\footnote{
  The instance sizes are bounded by the memory footprint of automatic differentiation through the MPS.
  During backpropagation, intermediate activations at each of the $N$ sites must be retained, leading to a temporary memory cost proportional to $B \times \chi^{2} \times N$, where $B$ is the training batch size and $\chi$ is the bond dimension.
  With $\chi = 128$ and $B = 2^{14}$, this easily exceeds $\mathcal{O}(1 \text{GB})$, limiting the full MPS to instance sizes $\sim 50$.
  The $k$-site variant alleviates this bottleneck by replacing $N$ with $k$ in the scaling, making it applicable to larger instances within the same memory budget.
  Minibatch training could further mitigate the memory constraint, but we do not pursue this direction in the current study for simplicity.
}.

For the softmax surrogate, we use rank-based weighting where tours are weighted by $\exp(-\mathrm{rank}/T)$ with temperature scaled by population size, and duplicate tours (ties) are removed before weighting.
The usual way to train MPS Born machines is a DMRG-based one~\cite{White:1992zz,White:1993zza,stoudenmire2016supervised,Han:2018jah}.
In this paper, however, we train Born machines by using the automatic differentiation technique for tensor networks~\cite{Liao:2019bye}.
The details of the training and a rough comparison with the DMRG-based approach are shown in appendix~\ref{sec:admps}.
Unless otherwise stated, we use the full ($k = N$) MPS Born machine for GEO.
If tour length is shown, it is normalized by the longest distance among the cities for each instance.

\subsection{Bond dimension dependence of distribution quality}

\begin{figure}[htbp]
  \centering
  \includegraphics[width=\hsize]{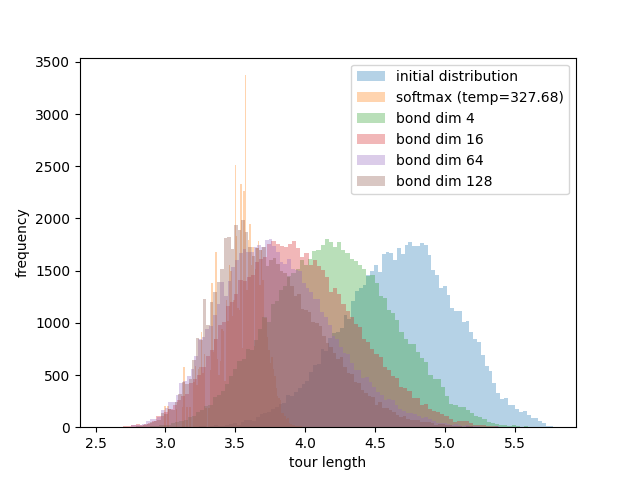}
  \caption{
    Effect of bond dimension on sampling quality for ulysses16.
    Tour length distributions are shown for the initial random population, the softmax-weighted target distribution, and MPS samples at bond dimensions $\in \{4, 16, 64, 128\}$.
    Higher bond dimensions enable the MPS to better approximate the target distribution.
    All distributions consist of $2^{16}$ samples.
  }
  \label{fig:bonddim_dependence}
\end{figure}

Before presenting optimization results, we examine how well the MPS Born machine learns a given softmax-weighted distribution.
Figure~\ref{fig:bonddim_dependence} shows the distribution of sampled tour lengths for different bond dimensions on the ulysses16 instance.
The initial random population shows a misplaced (larger cost) distribution, while the softmax-weighted target concentrates probability on shorter tours.
As bond dimension increases from 4 to 128, the MPS samples progressively better approximate the target distribution,
demonstrating that bond dimension controls the model's capacity to represent complex probability distributions over permutations.

This confirms that larger bond dimensions lead to higher-quality samples, which can be crucial for the optimization performance.
Note, however, that one does not need to strictly stick to the large-bond-dimension regime in TN-GEO in terms of balancing exploration and exploitation.
Of course, too poor a Born machine would yield a nearly uniform distribution,
so expressiveness of MPS is mandatory to some extent.
See also ref.~\cite{Gardiner:2024fzq} for the relationship between expressiveness and exploration ability.

\subsection{Optimization results on TSPLIB instances}

In this subsection, we show optimization results on TSPLIB instances.
For comparison, we include two classical hill-climbing heuristics: swap local search and 2-opt local search.
Both start from a sequential tour $(1, 2, \ldots, N)$ and terminate when no improving move exists.
For details, we provide pseudocodes of the algorithms in appendix~\ref{sec:heuristics}.
We choose these local search heuristics as baselines because they share a key structural feature with the MPS-based generative model:
they exploit local correlations among cities.
Swap modifies a tour by exchanging two city positions, while 2-opt reverses a contiguous segment;
both operate on short-range neighborhoods of the current tour.
Similarly, the MPS Born machine---and especially the $k$-site variant introduced in sec.~\ref{sec:k_site_geo}---captures correlations among consecutive cities in the tour sequence.
Comparing TN-GEO against these local-move heuristics therefore isolates the advantage of learning local structure from data via a generative model, as opposed to searching through local neighborhoods greedily.
Note, however, that these heuristics are not intended as competitive TSP solvers;
in practice, such local moves can be far more powerful when embedded within metaheuristic frameworks such as simulated annealing or multi-start methods.

We employ temperature scheduling with $T_{\mathrm{init}} = 0.1$ and $T_{\mathrm{final}} = 0.0001$ over the GEO iterations.
Note that this parameter is multiplied by the number of distinct samples at each iteration, as explained in sec.~\ref{sec:setup}.
Hereafter, the bond dimension of the MPS Born machine is set to 128 regardless of the problem sizes.
In any case, the optimization starts from $2^{16}$ random initial tours.
At each iteration, $2^{14}$ samples are drawn by the finite-temperature softmax surrogate for training, and $2^{16}$ new samples are generated from the trained MPS.

\begin{figure}[htbp]
  \centering
  \includegraphics[width=\hsize]{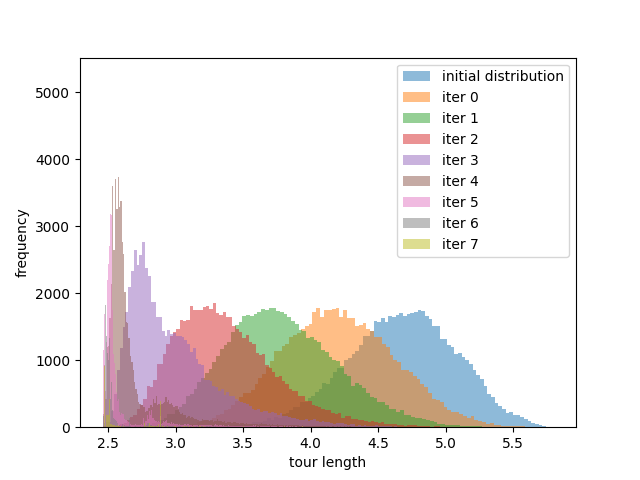}
  \caption{
    Evolution of tour length distribution across GEO iterations for ulysses16.
    The distribution progressively shifts toward shorter tour lengths as iterations proceed.
  }
  \label{fig:ulysses16_distribution}
\end{figure}

\begin{figure*}[t]
  \centering
  \includegraphics[width=\columnwidth]{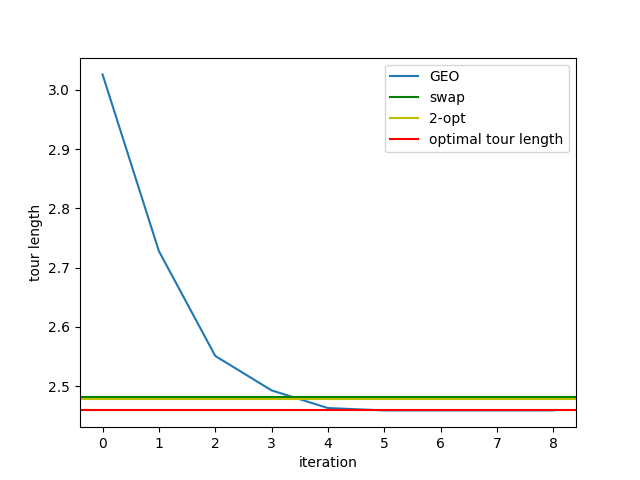}
  \includegraphics[width=\columnwidth]{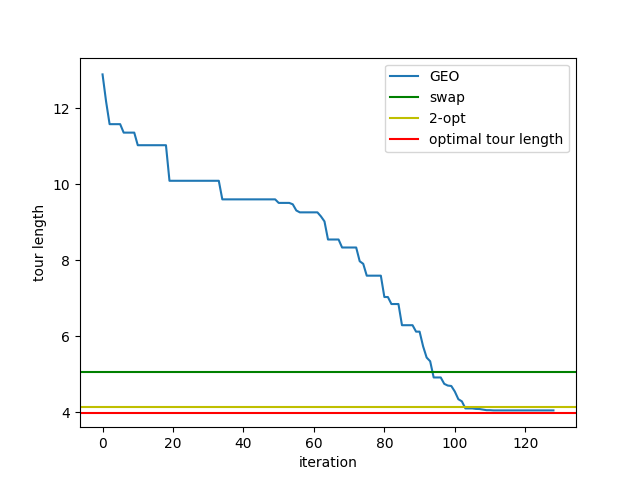}
  \caption{
    Convergence of best tour length for ulysses16 (left, 16 cities) and att48 (right, 48 cities).
    TN-GEO converges to or near the optimal solution, outperforming both swap and 2-opt heuristics.
  }
  \label{fig:history}
\end{figure*}

We first demonstrate TN-GEO on two representative instances: ulysses16 (16 cities) and att48 (48 cities).
Figure~\ref{fig:ulysses16_distribution} shows how the tour length distribution evolves across GEO iterations for ulysses16, progressively concentrating on shorter tours.
Figure~\ref{fig:history} compares the convergence of best tour length for both instances.
For ulysses16, TN-GEO reaches the known optimal solution within 8 iterations and outperforms the classical heuristics.
For att48, although the progressive improvement over iterations and superiority to the classical heuristics are common with the smaller ulysses16, it converges after a relatively large number of iterations.
Moreover, the att48 result is not exactly optimal.

\begin{figure*}[t]
  \centering
  \includegraphics[width=\columnwidth]{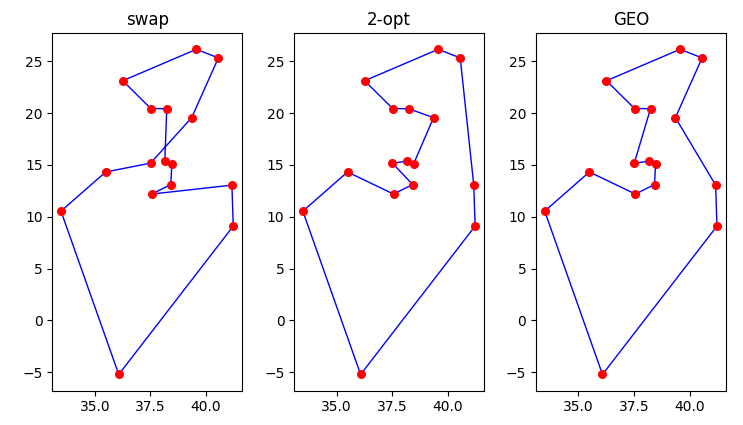}
  \includegraphics[width=\columnwidth]{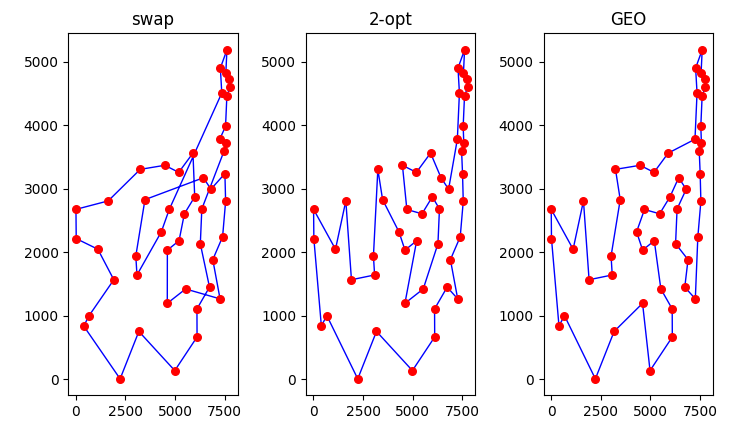}
  \caption{
    Best tours found for ulysses16 (left three) and att48 (right three).
    For comparison, the tours that are found by the swap and 2-opt heuristics are also shown.
  }
  \label{fig:tour}
\end{figure*}

Figure~\ref{fig:tour} visualizes the best tours found by TN-GEO.
We can observe that TN-GEO can eliminate tour crossings even when it cannot reach the optimum for att48.

Table~\ref{tab:results} summarizes the performance across various TSPLIB instances.
All values are reported as the gap to the known optimal tour length (\%), where lower is better.
In the table, we include both the full $N$-site MPS and the $k$-site variants with $k \in \{2, 4, 8\}$,
where the number of iterations is fixed to 64 across all instances~\footnote{
  Note that the performance of GEO can monotonically improve with an increasing number of iterations thanks to its reinforcement learning nature.
  In the table, however, we use the specific choice for simplicity.
}.
As shown in sec.~\ref{sec:k_site_geo}, the $k$-site variants are fed $N$ times more training samples compared to the full TN-GEO for a given distribution of tours.
An option is to reduce the number of tours for the $k$-site variants to match the volume of training data to that in the full TN-GEO.
In this section, however, we do not take such a prescription, to keep the diversity of training tours consistent between the models.

For small instances (burma14, ulysses16, ulysses22), all TN-GEO variants including the most compact $k=2$ model find the optimal tour.
For larger instances (att48, eil51, berlin52), while the $k=2$ and the full TN-GEO start struggling to converge to the optimal, the $k=4$ and $k=8$ TN-GEOs consistently achieve the optimal tour, outperforming both classical heuristics.
This demonstrates the superiority of the moderate $k$-site models to the too poor $k=2$ and the too rich (full) models:
given the limited bond dimensions, focusing only on $k$-grams is more efficient despite discarding the long-range relationships among cities owing to its reduced parameter space.
This result suggests, in a sense, that rough models perform exploration over the solution space better than well-tuned models~\cite{Gardiner:2024fzq}.

While the general phenomenon---that moderate expressiveness can outperform full expressiveness under limited bond dimension---is a property of the exploration-exploitation trade-off in GEO, the particular value of $k$ that performs best reflects the correlation structure of the problem.
For TSP, good tours are largely characterized by their local geometric structure: which cities should appear as consecutive neighbors.
The success of moderate $k$ indicates that learning good $k$-grams is sufficient to reconstruct near-optimal tours, consistent with the geometric locality of TSP.
For combinatorial optimization problems where solution quality depends on longer-range correlations, a larger $k$ would be expected to be necessary.

Notably, while the solution quality of swap and 2-opt heuristics gets almost monotonically worse with increasing instance size, the TN-GEO's results are relatively stable in this range.
Also, even for the larger half instances, TN-GEO performs comparably to or better than the 2-opt results.

\begin{table*}[htbp]
  \centering
  \caption{
    Gap to optimal tour length (\%) on TSPLIB instances.
    Best result per instance in bold.
    While the swap and the 2-opt results are deterministic, TN-GEO results can fluctuate per trial.
    In this table, however, we show only single-run results for TN-GEO.
    Values are shown to 3 significant figures.
  }
  \label{tab:results}
  \begin{tabular}{l|rr|rrrrrr|r}
    \hline
    Instance & Swap & 2-opt & $k=2$ & $k=4$ & $k=8$ & full \\
    \hline
    burma14 & 7.1\% & 3.76\% & \textbf{0\%} & \textbf{0\%} & \textbf{0\%} & \textbf{0\%} \\
    ulysses16 & 0.933\% & 0.787\% & \textbf{0\%} & \textbf{0\%} & \textbf{0\%} & \textbf{0\%} \\
    ulysses22 & 21.6\% & 1.47\% & \textbf{0\%} & \textbf{0\%} & \textbf{0\%} & \textbf{0\%} \\
    att48 & 26.7\% & 3.45\% & 6.67\% & \textbf{0\%} & \textbf{0\%} & 6.13\% \\
    eil51 & 30\% & 4.93\% & 3.05\% & \textbf{0\%} & \textbf{0\%} & 3.99\% \\
    berlin52 & 29.3\% & 16.5\% & 3.9\% & \textbf{0\%} & \textbf{0\%} & 4.18\% \\
    \hline
  \end{tabular}
\end{table*}

\section{Conclusion}
\label{sec:conclusion}

This study presents a tensor network Born machine approach for solving the TSP within the generator-enhanced optimization framework.
Key components are threefold.
First, we autoregressively sample permutations with masking:
by computing Born rule probabilities and masking previously selected cities, the MPS directly generates valid permutations without rejection sampling or post-selection.
Second, by utilizing the symmetry that the TSP has, we introduced the $k$-site MPS variant for TN-GEO, a parameter-efficient approach that learns distributions over $k$-grams using a sliding window, enabling scaling to larger instances by trading off model expressiveness for computational efficiency.
Third, we employed temperature scheduling: exponential temperature scheduling in the softmax surrogate helps GEO converge to a better point by balancing exploration and exploitation.
On the implementation side, the MPS is trained via automatic differentiation~\cite{Liao:2019bye} rather than the conventional DMRG-based approach, enabling global updates of all tensor elements through gradient-based optimization.

Experimental results on TSPLIB instances demonstrate that TN-GEO can outperform classical hill-climbing heuristics (swap and 2-opt).
In particular, moderate $k$-site models consistently outperform the full MPS, indicating that near-optimal tours are largely determined by local geometric correlations among consecutive cities---precisely the type of correlations that the MPS chain structure is well suited to capture.
Although our experiments are limited to instances with up to 52 cities, the solution quality of TN-GEO remains relatively stable across this range.

This finding is consistent with observations in other GEO applications:
Vodovozova \textit{et al.}~\cite{Vodovozova:2025idx} similarly found that reduced model expressiveness can improve GEO performance for the multi-knapsack problem, but also reported that MPS struggles at larger scales where long-range correlations become essential.
Our TSP results present the converse---because good tours are governed by local geometry, the limited correlation range of MPS is not a bottleneck but rather a useful inductive bias.
Together with the strong encoding dependence observed in production planning applications~\cite{Banner:2023abq}, these findings suggest that the effectiveness of MPS-based GEO hinges on the alignment between the problem's correlation structure and the MPS chain encoding.
It is worth noting that a primary motivation for quantum-inspired approaches in machine learning has been the entanglement structure of quantum states, which underpins the expressiveness of tensor network models.
Our results, together with those of ref.~\cite{Gardiner:2024fzq}, suggest a nuanced picture: within the GEO framework, restricting expressiveness can be beneficial for exploring the solution space, as an overly expressive model may concentrate too quickly on a narrow region.
This does not diminish the importance of entanglement-based expressiveness in general, but it does highlight that the interplay between model capacity and optimization dynamics deserves careful consideration.

This work highlights the versatility of tensor network methods for combinatorial optimization.
Compared to purely quantum Born machines, classical tensor networks allow flexible sampling schemes that can enforce hard constraints exactly, reducing or eliminating the reliance on penalty terms~\cite{Lucas:2013ahy}.

Future work may extend this approach to related scheduling and routing problems where tensor network methods have been analyzed~\cite{MataAli:2023onx}.
In the original paper~\cite{Alcazar:2021wjc}, it is claimed that one of the advantages of GEO is a reduced number of cost function evaluations that can be achieved by a weighting technique for training samples.
This feature is beneficial for problems where evaluation of the cost function is a bottleneck.
For TSP, however, the cost function evaluation is simply a summation of pairwise distances and thus inexpensive,
so evaluating every sample explicitly does not harm the optimization process.
Applying GEO to problems with expensive cost functions, where the evaluation-efficient nature of GEO can be fully leveraged, would be an interesting direction for future work.

\section*{Acknowledgments}

This work was performed for the Council for Science, Technology and Innovation (CSTI), Cross-ministerial Strategic Innovation Promotion Program (SIP), ``Promoting the application of advanced quantum technology platforms to social issues'' (Funding agency: QST).


\appendix

\section{Automatically Differentiable MPS Born Machine}
\label{sec:admps}

In this work, the MPS Born machine is implemented using JAX~\cite{jax} for automatic differentiation (AD), enabling gradient-based optimization of the MPS tensors without manually deriving gradients (see also ref.~\cite{Liao:2019bye}).

The MPS consists of $N$ tensors $\{A^{[k]}\}_{k=1}^N$ (or $k$ tensors for the $k$-site variant).
Each tensor $A^{[k]}$ has shape $(\chi, d_k, \chi)$, where $\chi$ is the bond dimension and $d_k$ is the physical dimension (number of cities, for TSP) at site $k$.
For boundary tensors, the external bond dimension is 1.
The tensors are initialized with random values.
In this work, we exclusively use real-valued tensors to construct MPS.
They are right-canonicalized upon calculating the negative log-likelihood (NLL) and sampling.

Given training data (tours weighted by the softmax surrogate), the model is trained by minimizing the NLL, as shown in algorithm~\ref{alg:admps}.

\begin{algorithm}[H]
  \caption{Automatically differentiable MPS training}
  \label{alg:admps}
  \begin{algorithmic}[1]
    \Require Training tours $\{\mathbf{p}_{(i)}\}$ with weights $\{Q_i\}$, learning rate $\eta$, tolerance $\epsilon$
    \Ensure Trained MPS tensors $\{A^{[k]}\}$
    \State Initialize MPS tensors $\{A^{[k]}\}$ randomly
    \State Initialize optimizer state
    \Repeat
      \State Right-canonicalize tensors via QR decomposition
      \State Compute NLL: $\mathcal{L} = -\sum_i Q_i \log P(\mathbf{p}_{(i)})$
      \State Compute gradient: $\nabla_{\{A^{[k]}\}} \mathcal{L}$ via AD
      \State Update tensors using AdamW: $A^{[k]} \gets A^{[k]} - \eta \cdot \mathrm{AdamW}(\nabla)$
    \Until{NLL improvement $< \epsilon$ for consecutive iterations}
    \State \Return $\{A^{[k]}\}$
  \end{algorithmic}
\end{algorithm}

The gradient is computed automatically using JAX's \texttt{grad} function, and we use the AdamW optimizer~\cite{loshchilov2018decoupled} in Optax~\cite{deepmind} with learning rate $\eta = 0.1$.
Throughout this paper, we set the convergence tolerance to $\epsilon = 10^{-4}$.

Sampling follows algorithm~\ref{alg:sampling}, proceeding autoregressively from left to right.
At each position, the left environment is contracted with the current tensor, Born rule probabilities are computed while masking previously selected cities, and the new city is sampled.

We compare this AD training approach with DMRG-style local optimization, which sweeps through pairs of adjacent tensors and performs updates via simple gradient descent of the merged tensor plus singular value decomposition (SVD) to retain the MPS structure.

\begin{figure}[htbp]
  \centering
  \includegraphics[width=\hsize]{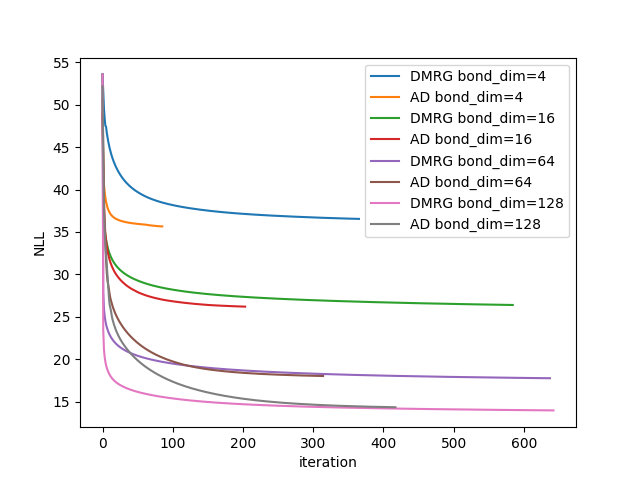}
  \caption{
    Comparison of NLL convergence between DMRG and AD training on ulysses16 for different bond dimensions.
    Both methods achieve similar final NLL values despite their different optimization strategies.
    Convergence tolerance is $10^{-4}$.
  }
  \label{fig:dmrg_vs_ad_nll}
\end{figure}

\begin{figure}[htbp]
  \centering
  \includegraphics[width=\hsize]{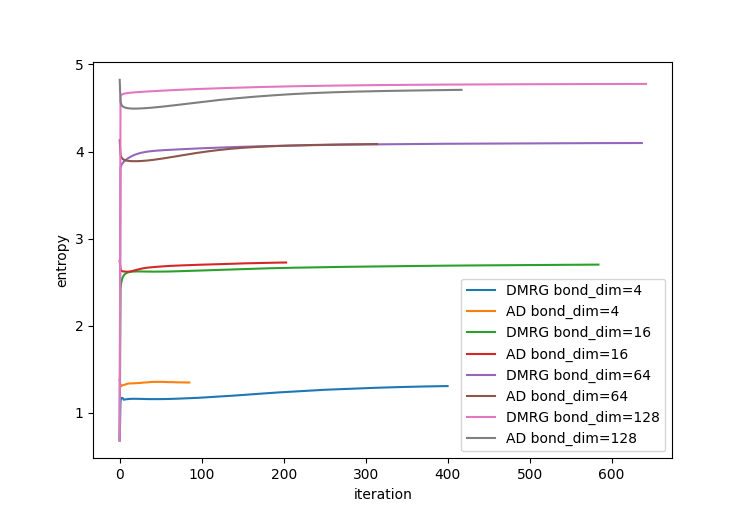}
  \caption{
    Bipartite entanglement entropy during training for DMRG and AD methods.
    Both methods converge to similar entropy values, indicating comparable quantum state structure.
    Convergence tolerance is $10^{-4}$.
  }
  \label{fig:dmrg_vs_ad_entropy}
\end{figure}

Figures~\ref{fig:dmrg_vs_ad_nll} and~\ref{fig:dmrg_vs_ad_entropy} show the NLL and bipartite entanglement entropy trajectories during training on ulysses16.
For the training, we have prepared $2^{16}$ random TSP tours and selected them with replacement through the finite-temperature softmax surrogate, as in the first step of the GEO workflow.
On the DMRG side, we count an iteration by consecutive right and left sweeps,
and, to make it consistent with DMRG as much as possible, we define an iteration of AD update by two global updates.

The bipartite entanglement entropy is defined by
\begin{align}
  S = -\sum_i p_i \log p_i,
\end{align}
where
\begin{align}
  p_i = \frac{s_i^2}{\sum_j s_j^2}
\end{align}
and where $\{s_i\}$ are the singular values obtained from SVD of the merged tensor $M_{(i,j),(k,l)} = \sum_{a} A^{[N/2-1]}_{i,j,a} A^{[N/2]}_{a,k,l}$ at the middle cut.
This quantity characterizes the entanglement structure of the learned quantum state.

Although the optimization strategies differ fundamentally---DMRG performs local updates with SVD truncation while AD globally updates the tensor elements using gradients---both methods achieve similar final NLL values and entropy profiles.
The convergence to similar entropy values suggests that both methods find MPS representations with comparable correlation structure, despite taking different optimization paths.

In the main body of this paper, we entirely use the AD approach expecting that the global update strategy is more suitable for our case, where the target distribution dynamically changes as the iteration step proceeds.
The AD approach has an advantage in terms of computational complexity too,
since the complexity of backpropagation does not exceed that of forward calculation.
In other words, the complexity of the AD update is determined by consecutive QR decompositions to make the MPS right-canonical and by the calculation of NLL.
This is less expensive than the DMRG update that involves site-by-site SVDs.

Note that these comparisons are not strict benchmarking.
Since the optimization policies are drastically different from each other,
the meaning of \textit{e.g.} learning rate cannot be the same between the two methods.
Indeed, the speed of convergence in terms of the number of iterations is not monotonic;
they just depend on the situation.

\section{Hill-Climbing Heuristics}
\label{sec:heuristics}

In the main body of the paper, we compare TN-GEO against two classical hill-climbing heuristics for TSP.
To ensure reproducibility, in this section we describe the exact procedure for both methods.

The first one is swap local search, which starts from a sequential tour and iteratively improves it by swapping pairs of cities, as shown in algorithm~\ref{alg:swap}.

\begin{algorithm}[H]
  \caption{Swap local search}
  \label{alg:swap}
  \begin{algorithmic}[1]
    \Require Distance matrix $d$, number of cities $N$
    \Ensure Locally optimal tour
    \State $\mathbf{x} \gets [0, 1, 2, \ldots, N-1]$
    \State $L^{*} \gets$ tour length of $\mathbf{x}$
    \Repeat
      \State $\textit{improved} \gets \textsc{False}$
      \For{$i = 0$ to $N-2$}
        \For{$j = i+1$ to $N-1$}
          \State $\mathbf{x}^{\prime} \gets \mathbf{x}$ with cities at positions $i$ and $j$ swapped
          \State $L^{\prime} \gets$ tour length of $\mathbf{x}^{\prime}$
          \If{$L^{\prime} < L^{*}$}
            \State $\mathbf{x} \gets \mathbf{x}^{\prime}$, $L^{*} \gets L^{\prime}$
            \State $\textit{improved} \gets \textsc{True}$
            \State \textbf{break} both loops
          \EndIf
        \EndFor
      \EndFor
    \Until{$\textit{improved} = \textsc{False}$}
    \State \Return $\mathbf{x}$
  \end{algorithmic}
\end{algorithm}

This is a first-improvement strategy; the search restarts immediately upon finding an improving move, converging to a local optimum where no single swap can improve the solution.

The second heuristic is 2-opt local search, which also starts from a sequential tour but improves it by reversing segments rather than swapping individual cities, as shown in algorithm~\ref{alg:twoopt}.

\begin{algorithm}[H]
  \caption{2-opt local search}
  \label{alg:twoopt}
  \begin{algorithmic}[1]
    \Require Distance matrix $d$, number of cities $N$
    \Ensure Locally optimal tour
    \State $\mathbf{x} \gets [0, 1, 2, \ldots, N-1]$
    \State $L^{*} \gets$ tour length of $\mathbf{x}$
    \Repeat
      \State $\textit{improved} \gets \textsc{False}$
      \For{$i = 0$ to $N-3$}
        \For{$j = i+2$ to $N-1$}
          \If{$i = 0$ and $j = N-1$}
            \State \textbf{continue} \Comment{Skip trivial reversal}
          \EndIf
          \State $\mathbf{x}^{\prime} \gets \mathbf{x}$ with segment $[i+1, j]$ reversed
          \State $L^{\prime} \gets$ tour length of $\mathbf{x}^{\prime}$
          \If{$L^{\prime} < L^{*}$}
            \State $\mathbf{x} \gets \mathbf{x}^{\prime}$, $L^{*} \gets L^{\prime}$
            \State $\textit{improved} \gets \textsc{True}$
            \State \textbf{break} both loops
          \EndIf
        \EndFor
      \EndFor
    \Until{$\textit{improved} = \textsc{False}$}
    \State \Return $\mathbf{x}$
  \end{algorithmic}
\end{algorithm}

The 2-opt move removes edges $(x_i, x_{i+1})$ and $(x_j, x_{j+1})$ and adds edges $(x_i, x_j)$ and $(x_{i+1}, x_{j+1})$.
This is particularly effective at removing edge crossings, as a reversal can untangle crossed edges.
The algorithm converges to a 2-optimal tour where no segment reversal can improve the solution.

Both heuristics are deterministic and fast, but they are prone to local optima.
Their solutions serve as baselines for comparing TN-GEO's ability to find globally better solutions.



%

\end{document}